\documentclass{article}
\usepackage{arxiv}

\usepackage[utf8]{inputenc} 
\usepackage[T1]{fontenc}    
\usepackage{hyperref}       
\usepackage{url}            
\usepackage{booktabs}       
\usepackage{amsfonts}       
\usepackage{nicefrac}       
\usepackage{microtype}      
\usepackage{graphicx}
\usepackage{caption}
\usepackage{subcaption}
\usepackage{url}
\usepackage{paralist} 
\usepackage{listings} 
\usepackage{color}
\usepackage{bm}
\usepackage{esvect}
\usepackage{amssymb}
\usepackage{float}
\restylefloat{table}
\usepackage[intlimits]{amsmath}
\usepackage{placeins}
\usepackage{chngpage}
\usepackage{sidecap}
\usepackage{flushend}

\allowdisplaybreaks

\usepackage{lipsum}
\usepackage{titlesec}

\titlespacing*{\section}{0pt}{7pt plus 0pt minus 0pt}{5pt plus 0pt minus 0pt}
\titlespacing*{\subsection}{0pt}{7pt plus 0pt minus 0pt}{5pt plus 0pt minus 0pt}
\newcommand{\modelname}{LaMP}
\newcommand{\fullmodelname}{Label Message Passing}

\newcommand{\mpnnxx}{MPNN$_{\textrm{xx}}$}
\newcommand{\mpnnxy}{MPNN$_{\textrm{xy}}$}
\newcommand{\mpnnyy}{MPNN$_{\textrm{yy}}$}

\newcommand{\gxx}{$G_{\textrm{xx}}$}
\newcommand{\gyy}{$G_{\textrm{yy}}$}

\usepackage{color}

\date{}

\title{Neural Message Passing for Multi-Label Classification}

\author{
  Jack Lanchantin, Arshdeep Sekhon, \& Yanjun Qi\\
  Department of Computer Science\\
  University of Virginia\\
  Charlottesville VA 22903 \\
  \texttt{\{jjl5sw,as5cu,yq2h\}@virginia.edu} \\
}

\begin{document}
\maketitle

\begin{abstract}

Multi-label classification (MLC) is the task of assigning a set of target labels for a given sample. Modeling the combinatorial label interactions in MLC has been a long-haul challenge. 
We propose \fullmodelname{} (\modelname{}) Neural Networks to efficiently model the joint prediction of multiple labels. \modelname{} treats labels as nodes on a 
label-interaction graph and computes the hidden representation of each label node conditioned on the input using attention-based neural message passing. Attention enables \modelname{} to assign different importances to neighbor nodes per label, learning how labels interact (implicitly). 
The proposed models are simple, accurate, interpretable,  structure-agnostic, and applicable for predicting dense labels since \modelname{} is incredibly parallelizable. 
We validate the benefits of \modelname{}  on seven real-world MLC datasets, covering a broad spectrum of input/output types and outperforming the state-of-the-art results. Notably, \modelname{} enables intuitive interpretation of how classifying each label depends on the elements of a sample and at the same time rely on its interaction with other labels\footnote{We provide our code and datasets at https://github.com/QData/LaMP}. 

\end{abstract}

\section{Introduction}
Multi-label classification (MLC) is receiving increasing attention in areas such as natural language processing, computational biology, and image recognition. Accurate and scalable MLC methods are in urgent need for applications like assigning topics to web articles, or identifying binding proteins on DNA. The most common and straightforward MLC method is the binary relevance (BR) approach that considers multiple target labels independently \cite{tsoumakas2006multi}. However, in many MLC tasks there is a clear dependency structure among labels, which BR methods ignore. 

Unfortunately, accurately modelling all combinatorial label interactions is an NP-hard problem. Many types of models, including a few deep neural network (DNN) based, have been introduced to approximately model such interactions, thus boosting classification accuracy. 

Our main concern of this paper is how to represent multiple labels jointly (and conditioned on the input features) in order to make accurate predictions. The most relevant literature addressing this concern falls roughly into three groups.

The first group, probabilistic classifier chain (PCC) models, formulate the joint label dependencies using the chain rule and perform MLC in a sequential prediction manner \cite{read2009classifier,wang2016cnn,nam2017maximizing}. 
Notably, \cite{nam2017maximizing} used a recurrent neural network (RNN) sequence to sequence (Seq2Seq) architecture \cite{graves2013generating} for MLC and achieved the state-of-the-art performance on multiple text-based datasets. 
However, these methods are inherently unfit for MLC tasks due to their incapacity to be parallelized, and inability to perform well in dense label settings, or when there are a large number of positive labels (since errors propagate in the sequential prediction). We refer the reader to the supplementary material for a full background and analysis of PCC methods (Appendix section \ref{sec:background}).
The second group learns a shared latent space representing both input features and output labels, and then upsamples from the space to reconstruct the target labels \cite{yeh2017learning,bhatia2015sparse}. The main drawback of this group is the interpretability issue with a learned low dimensional latent space, as many real-world applications prefer interpretable predictors. 
The third group models conditional label dependencies using a structured output or graphical model representation \cite{lafferty2001conditional,tsochantaridis2005large}. However, these methods are often limited to only considering pair-wise dependencies due to computational constraints, or are forced to use some variation of approximate inference which has no clear representation of conditional dependencies.

Thus our main question is: \emph{is it possible to have accurate, flexible and explainable MLC methods that are applicable to many dense labels?} This paper provides empirical results showing that this is possible through extending attention based Message Passing Neural Networks (MPNNs) to learn the joint representation of multiple labels conditioned on input features.

MPNNs \cite{gilmer2017neural} are a class of methods that efficiently learn the joint representations of variables using neural message passing strategies. They provide a flexible framework for modeling multiple variables jointly which have no explicit ordering.  

The key idea of our method is to rely on attention-based neural message passing entirely to draw global dependencies from labels to input features, and from labels to labels. To the best of our knowledge, this is the first extension of MPNNs to model a conditional joint representation of output labels, and additionally the first extension of MPNNs to model the interactions of variables where the exact structure is unknown. We name the proposed method \fullmodelname{} (\modelname{}) Networks since it performs neural message passing on an unknown, fully-connected label-to-label graph.  Through  intra-attention (aka self-attention),  \modelname{}  assigns different importance to different neighbor nodes per label, dynamically learning how labels interact conditioned on a specific input. We further extend \modelname{} to cases when a known label interaction graph is provided by modifying the intra-attention to only attend over a node's known neighbors. \modelname{} networks allow for parallelization in training and testing and can work with dense labels, overcoming the drawbacks of PCC methods.

\modelname{} most closely belongs to the third MLC category we mentioned above, however it trains a unified model to classify each label \emph{and} model the label to label dependencies at the same time, in an end-to-end fashion. The important aspect is that \modelname{} networks automatically learn the output label dependency structure conditioned on a specific input using neural message passing. This in turn can easily be interpreted to understand the conditional structure. 

The main contributions of this paper include: (1) \textbf{Accurate MLC}: Our model achieves similar, or better performance compared to the previous state of the art across five MLC metrics. We validate our model on eight MLC datasets which cover a wide spectrum of input data structure: sequences (English text, DNA), tabular (binary word vectors), graph (drug molecules), and images, as well as output label structure:  unknown and graph. (2) \textbf{Interpretable}: Although deep-learning based systems have widely been viewed as ``black boxes'', our attention based \modelname{} models allow for a straightforward way to extract three different types of model visualization: intermediate network predictions, label to feature dependencies, and label to label dependencies.

\setlength{\abovedisplayskip}{6pt}
\setlength{\belowdisplayskip}{6pt}

\section{Method: \modelname{} Networks}
\label{sec:method}

\textbf{Notations.} We define the following notations, used throughout the paper. Let $\mathcal{D} = \{(\bm{x}_n,\bm{y}_n)\}_{n=1}^{N}$ be the set of data samples with inputs $\bm{x} \in X$ and outputs $\bm{y} \in Y$. Inputs $\bm{x}$ are a (possibly ordered) set of $S$ components $\{{x}_1,{x}_2,...,{x}_S\}$, and outputs $\bm{y}$ are a set of $L$ labels $\{{y}_1,{y}_2,...,{y}_L\}$. MLC involves predicting the set of binary labels $\{{y}_1,{y}_2,...,{y}_L\}, y_i \in \{0,1\}$ given input $\bm{x}$.

In general we can assume to represent the input feature components as embedded vectors $\{\bm{z}_1,\bm{z}_2,...,\bm{z}_S\}$, $\bm{z}_i \in \mathbb{R}^{d}$, using some learned embedding matrix $\mathbf{W}^x \in \mathbb{R}^{\delta \times d}$. Here  $d$ is the embedding size and, $\delta$ is the  size of ${x}_i$. ${x}_i$ can be any component of a particular input (for example, words in a sentence, patches of an image, nodes of a known graph, or one of the tabular features).

Similarly, labels can be first represented as embedded vectors $\{\bm{u}^{t=0}_1,\bm{u}^{t=0}_2,...,$ $\bm{u}^{t=0}_L\}$, $\bm{u}^t_i \in \mathbb{R}^{d}$,  through a learned embedding matrix $\mathbf{W}^y \in \mathbb{R}^{L \times d}$, where $L$ denotes the number of labels. 
Here we use $t$ to represent the `state' of the embedding after the $t^{th}$ update step. This is because in \modelname{} networks, each label embedding is updated for $t$ steps before the predictions are made. The key idea of \modelname{} networks is that labels are represented as nodes in a label-interaction graph \gyy{} denoting nodes as embedding vectors $\{\bm{u}^t_{1:L}\}$. \modelname{} networks use MPNN modules with attention to pass messages  from input embeddings $\{\bm{z}_{1:S}\}$ to \gyy{}, and then within \gyy{} to model the joint prediction of labels.

\vspace{-2mm}
\subsection{Background: Message Passing Neural Networks}

Message Passing Neural Networks (MPNNs) \cite{gilmer2017neural} are a generalization of graph neural networks (GNNs) \cite{scarselli2009graph}.
MPNNs model variables as nodes on a graph $G$. Here $G=(V,E)$, where $V$ describes the set of nodes (variables) and $E$ denotes the set of edges (about how variables interact with other variables). In MPNNs, joint representations of nodes and edges are modelled using message passing rather than explicit probabilistic formulations, allowing for efficient inference. MPNNs model the joint dependencies using message function $M^t$ and node update function $U^t$ for $T$ time steps, where $t$ is the current time step. The hidden state $\bm{v}^t_i \in \mathbb{R}^{d}$ of node $i \in G$ is updated based on messages $\bm{m}^{t}_i$ from its neighboring nodes $\{\bm{v}^t_{j \in \mathcal{N}(i)}\}$ defined by neighborhood $\mathcal{N}(i)$:
\begin{align}
    \bm{m}^{t}_i &= \sum_{j\in \mathcal{N}(i)} M^t(\bm{v}^t_i,\bm{v}^t_j),\\
    \bm{v}^{t+1}_i &= U^t(\bm{m}^{t}_i ).
\end{align}
After $T$ rounds of iterative updates to spread information to distant nodes, a readout function $R$ is used on the updated node embeddings to make  predictions like classifying nodes or classifying properties about the  graph. 



Many possibilities exist for functions $M^t$ and $U^t$. We specifically choose to pass messages using 
intra-attention (also called as self-attention) neural message passing which enable nodes to attend over their neighborhoods differentially. This allows for the network to learn different importances for different nodes in a neighborhood, without depending on knowing the graph structure upfront (essentially learning the unknown graph structure) \cite{velivckovic2017graph}. In this formulation, messages for node $\bm{v}^t_i$ are obtained by a weighted sum of all its neighboring nodes $\{\bm{v}^t_{j\in \mathcal{N}(i)}\}$ where the weights are calculated by attention representing the importance of each neighbor for a specific node~\cite{bahdanau2014neural}. In the rest of the paper, we use ``graph attention'' and ``neural message passing'' interchangeably.

Intra-attention neural message passing works as follows. We first calculate attention weights $\alpha^t_{ij}$ for pair of nodes ($\bm{v}^t_i$, $\bm{v}^t_j$) using attention function $a(\cdot)$:
\begin{align}
    \alpha^t_{ij} = \textrm{softmax}_j(e^t_{ij}) &=  \frac{\textrm{exp}(e^t_{ij})}{\sum_{k \in \mathcal{N}(i)}{\textrm{exp}(e^t_{ik})}} \label{eq:alpha} \\
    e^t_{ij} &= a(\bm{v}^t_i,\bm{v}^t_j) \label{labelnodeimportance} \\
        a(\bm{v}^t_i,\bm{v}^t_j) &= \frac{(\mathbf{W}^q\bm{v}^t_i)^{\top} (\mathbf{W}^u\bm{v}^t_j)}{\sqrt{\smash[b]d}}    
\label{labelnodea} 
\end{align}
where $e^t_{ij}$ represents the importance of node $j$ for node $i$, however un-normalized.  $e^t_{ij}$ are normalized across all neighboring nodes of node $i$ using a softmax function (Eq~\ref{eq:alpha}) to get $\alpha^t_{ij}$. For the attention function $a(\cdot)$, we used a scaled dot product with node-wise linear transformations $\mathbf{W}^q \in \mathbb{R}^{d \times d}$ on node $\bm{v}^t_i$ and $\mathbf{W}^u \in \mathbb{R}^{d \times d}$ on node $\bm{v}^t_j$. Scaling by $\sqrt{\smash[b] d}$ is used to mitigate training issues \cite{vaswani2017attention}.

Then we use a so called attention message function $M_{\textrm{atn}}^t$ to produce the message from node $j$ to node $i$ using the learned attention weights $\alpha^t_{ij}$ and another transformation matrix $\mathbf{W}^v \in \mathbb{R}^{d \times d}$:
\begin{gather}
    \label{eq:m_attention}
    M_{\textrm{atn}}(\bm{v}^t_i,\bm{v}^t_j; \bm{W}) = \alpha^t_{ij} \mathbf{W}^v \bm{v}^t_j,\\
    \bm{m}^t_i = \bm{v}^t_i + 
    \sum_{j\in \mathcal{N}(i)} M_{\textrm{atn}}(\bm{v}^t_i,\bm{v}^t_j; \bm{W}). \label{eq:mt}
\end{gather}
Eq~\ref{eq:mt} computes the full message $\bm{m}^t_i$ for node $\bm{v}^t_i$ by linearly combining messages from all neighbor nodes $j \in \mathcal{N}(i)$ with a residual connection on the current $\bm{v}^t_i$. 

Lastly, node $\bm{v}^t_i$ is updated to next state $\bm{v}^{t+1}_i$ using message $\bm{m}^t_i$ by a multi-layer perceptron (MLP) update function $U_{\textrm{mlp}}$, plus a $\bm{m}^t_i$ residual connection:
\begin{align}
    U_{\textrm{mlp}}(\bm{m}^{t}_i; \bm{W}) &= \textrm{ReLU}(\mathbf{W}^{r}\bm{m}^t_i + b_1)^{\top}\mathbf{W}^{b} + b_2\\
    {\bm{v}}^{t+1}_i &= \bm{m}^t_i + U_{\textrm{mlp}}(\bm{m}^{t}_i; \bm{W}).
    \label{eq:u_attention}
\end{align}
Function $U_{\textrm{mlp}}$ is parameterized  with matrices $\{\mathbf{W}^{r} \in \mathbb{R}^{d \times d}, \mathbf{W}^{b} \in \mathbb{R}^{d \times d}\}$. It is important to note that $\mathbf{W}$ in Eq~\ref{eq:u_attention} are shared (i.e., separately applied) across all nodes. This can be viewed as 1-dimensional convolution operation with kernel  and stride sizes of 1. Weight sharing across nodes is a key aspect of MPNNs, where node dependencies are learned in an order-invariant manner.


\subsection{\modelname{}: \fullmodelname{} } 

Given the input embeddings $\{\bm{z}_1,\bm{z}_2,...,\bm{z}_S\}$, the goal of \fullmodelname{} is to model the conditional dependencies between label embeddings $\{\bm{u}^{t}_1,\bm{u}^{t}_2,...,\bm{u}^{t}_L\}$ using Message Passing Neural Networks.
We assume that the label embeddings are nodes on a label-interaction graph called \gyy{}, where the initial state of the embeddings $\{\bm{u}^0_{1:L}\}$ at $t=0$ are obtained using label embedding matrix $\mathbf{W}^y$.

\begin{figure}[h]
    \vspace{-10pt}
    \centering
    \includegraphics[scale=0.6]{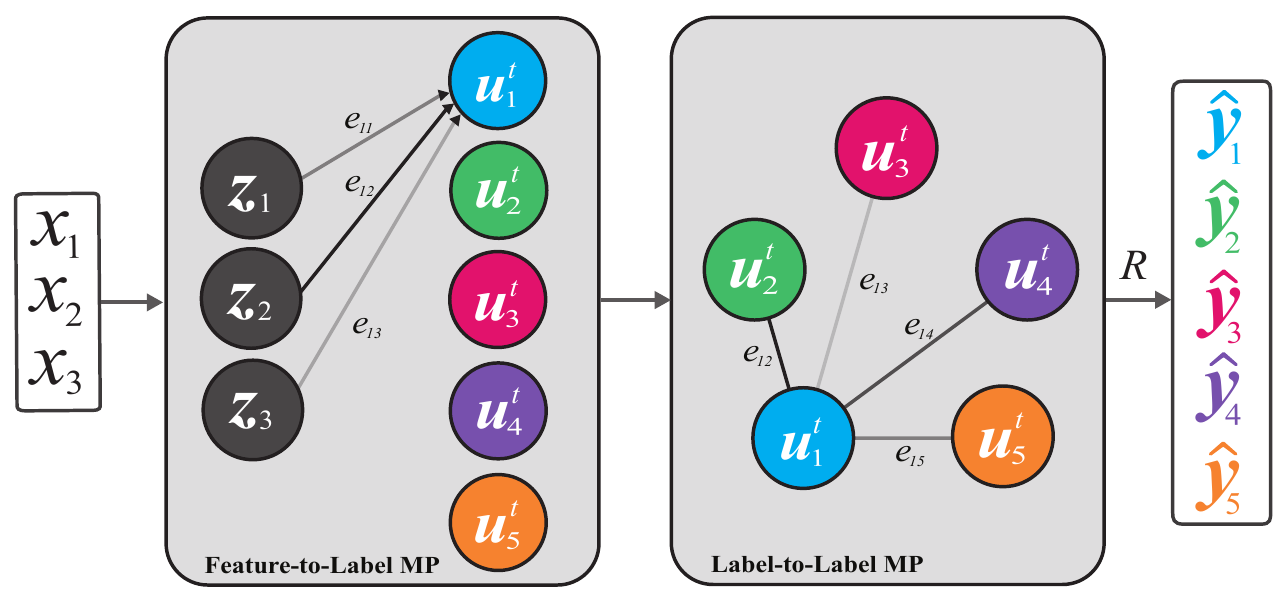}
     \caption{
    \footnotesize \textbf{\modelname{} Networks.} 
        Given input $\mathbf{x}$, we encode its components $\{x_1,x_2,x_3\}$ as embedded input nodes $\{\bm{z}_1,\bm{z}_2,\bm{z}_3\}$. 
        We encode labels $\{y_1,y_2,...,y_5\}$ as embedded label nodes
        $\{\bm{u}^t_1,\bm{u}^t_2,...,\bm{u}^t_5\}$ of label-interaction graph \gyy{}.  
        First, \mpnnxy{} is used to pass messages from the input nodes to the labels nodes and update the label nodes. Then, \mpnnyy{} is used to pass messages between the label nodes and update label nodes.  Finally, readout function $R$ performs node-level classification on label nodes to make binary label predictions $\{\hat{y}_1,\hat{y}_2,...,\hat{y}_5\}$. 
     }
    \label{fig:model}
\end{figure}

Each step $t$ in \fullmodelname{} consists of two parts in order to update the label embeddings: (a). Feature-to-Label Message Passing, where messages are passed from the input embeddings to the label embeddings, and (b). Label-to-Label Message Passing, where messages are passed between labels. An overview of our model is shown in Fig. \ref{fig:model}. We explain these two parts in detail in the following subsections. \modelname{} Networks use $T$ steps of attention-based neural message passing to update the label nodes before a readout function makes a prediction for each label $i$ on its final state $\bm{u}_i^T$.

\noindent \textbf{Updating Label Embeddings via Feature-to-Label Message Passing} 
\vspace{2pt}

\noindent Given a particular input $\bm{x}$ with embedded feature components $\{\bm{z}_1,\bm{z}_2,...,\bm{z}_S\}$, the first step in \modelname{} is to update the label embeddings by passing messages from the input embeddings to the label embeddings, as shown in the ``Feature-to-Label MP'' block of Fig. \ref{fig:model}.
To do this, \modelname{} uses neural message passing module \mpnnxy{} to update the $i^{th}$ label node's embedding $\bm{u}^t_i$ using the embeddings of all the components of an input.

That is, we update each $\bm{u}^t_i$ by using a weighted sum of all input embeddings $\{\bm{z}_{1:S}\}$, in which the weights represent how important an input component is to the $i^{th}$ label node. The weights for the message are learned via Label-to-Feature attention (i.e., each label attends to each input embedding differently to compute the weights). 

In this step, messages are only passed from the input nodes to the label nodes, and not vice versa (i.e. Feature-to-Label message passing is directed). 

More specifically, to update label embedding $\bm{u}^t_i$, \mpnnxy{} uses attention message function $M_{\textrm{atn}}^t$ on all embeddings of the input $\{\bm{z}_{1:S}\}$ to produce messages $\bm{m}^t_i$, and MLP update function $U_{\textrm{mlp}}$ to produce the updated intermediate embedding state $\bm{u}^{t'}_i$:
\vspace{-10pt}
\begin{gather}
    \bm{m}^t_i = \bm{u}^t_i + 
    \sum\limits_{j=1}^S M_{\textrm{atn}}(\bm{u}^t_i,\bm{z}_j; \bm{W}_{\textrm{xy}}), \\
    \bm{u}^{t'}_i = \bm{m}^t_i + U_{\textrm{mlp}}(\bm{m}^{t}_i; \bm{W}_{\textrm{xy}}).
\end{gather}
The key advantage of Feature-to-Label message passing with attention is that each label node can attend  differently on input elements (e.g. different words in an input sentence).

\vspace{2mm}
\noindent \textbf{Updating Label Embeddings via Label-to-Label Message Passing}
\vspace{2pt}

\noindent At this point, an independent prediction can be made for each label conditioned on $\bm{x}$ using $\{\bm{u}^{t'}_{1:L}\}$. However, in order to consider label dependencies, we model interactions between the label nodes $\{\bm{u}^{t'}_{1:L}\}$ using Label-to-Label message passing and update them accordingly, as shown in the ``Label-to-Label MP'' block of Fig. \ref{fig:model}. Given the exponentially large number of possible conditional dependencies, we use neural message passing as an efficient way to much such interactions, which has been shown to work well in practice for other tasks. 

We assume there exist a label interaction graph \gyy{} $ = (V_{\textrm{yy}},E_{\textrm{yy}})$, $V_{\textrm{yy}}=\{y_{1:L}\}$, and $E_{\textrm{yy}}$ includes all undirected pairwise edges connecting node $\bm{y}_i$ and node $\bm{y}_j$.
At this stage, we use another message passing module, \mpnnyy{} to pass messages between labels and update them. The label embedding $\bm{u}^{t'}_{i}$ is updated by a weighted combination through attention of all its neighbor label nodes $\{\bm{u}^{t'}_{j \in \mathcal{N}(i)}\}$.

\mpnnyy{} uses attention message function $M_{\textrm{atn}}^{t'}$ on all neighbor label embeddings $\{\bm{u}^{t'}_{j \in \mathcal{N}(i)}\}$ to produce message $\bm{m}^t_i$, and MLP update function $U_{\textrm{mlp}}^{t'}$ to compute updated embedding $\bm{u}^{t+1}_i$:
\begin{gather}
    \label{eq:label_to_label_M}
    \bm{m}^{t'}_i = \bm{u}^{t'}_i + 
    \sum_{j \in \mathcal{N}(i)} M_{\textrm{atn}}(\bm{u}^{t'}_i,\bm{u}^{t'}_j; \bm{W}_{\textrm{yy}}), \\
    {\bm{u}}^{t+1}_i = \bm{m}^{t'}_i + U_{\textrm{mlp}}(\bm{u}^{t'}_i , \bm{m}^{t'}_i; \bm{W}_{\textrm{yy}}).
\end{gather}

If there exists a known label interaction graph \gyy{}, message $\bm{m}^t_i$ for node $i$ is computed using its neighboring nodes $j \in \mathcal{N}(i)$, where the neighbors $\mathcal{N}(i)$ are defined by the graph. If there is no known \gyy{} graph, we assume a fully connected graph, which means $\mathcal{N}(i) =  \{j \in V_{\textrm{yy}}\}$ (including $i$).

\vspace{2pt plus 0pt minus 0pt}
\noindent \textbf{Message Passing for Multiple Time Steps} 

\noindent To learn more complex relations among nodes, we compute a total of $T$ time steps of updates. This is essentially a stack of $T$ MPNN layers. In our implementation, the label embeddings are updated by \mpnnxy{} and \mpnnyy{} for $T$ time steps to produce $\{\bm{u}^T_1,\bm{u}^T_2,...,\bm{u}^T_L\}$.

\subsection{Readout Layer (MLC Predictions from the label embeddings)} After $T$ updates to the label embeddings, the last module predicts each label $\{\hat{y}_1,...\hat{y}_L$\}. A readout function $R$ projects each of the $L$ label embeddings $\bm{u}^T_i$ using projection matrix $\mathbf{W}^o \in \mathbb{R}^{L\times d}$, where row $\mathbf{W}_i^o \in \mathbb{R}^{d}$ is the learned output vector for label $i$. The calculated vector of size $L\times1$ is then fed through an element-wise sigmoid function to produce probabilities of each label being positive:
\begin{equation}
   \label{eq:readout}
    \hat{y}_i = R({\bm{u}}^T_i;{\mathbf{W}^o}) = \textrm{sigmoid}({\mathbf{W}^o_i} {\bm{u}}^T_i).
\end{equation} 

\subsection{Model Details}
\label{sec:layer_details}

\textbf{Multi-head Attention.\,} In order to allow a particular 
node to attend to multiple other nodes (or multiple groups of nodes) at 
once, \modelname{} uses multiple attention heads. Inspired by 
\cite{vaswani2017attention}, we use $K$ independent attention heads for 
each $\mathbf{W}^{\cdot}$ matrix during the message computation, where 
each matrix column $\mathbf{W}^{\cdot,k}_j$ is of dimension $d/K$. The 
generated representations are concatenated (denoted by $\|$) and 
linearly transformed by matrix $\mathbf{W}^z \in \mathbb{R}^{d\times 
d}$. Multi-head attention changes message passing function 
$M_{\textrm{atn}}$, but update function $U_{\textrm{mlp}}$ stays the 
same.
\begin{gather}
    e^{t,k}_{ij} = (\mathbf{W}^{q,k}\bm{v}^t_i )^\top(\mathbf{W}^{u,k}\bm{v}^t_j) / \sqrt{d}\\
    \alpha^{t,k}_{ij} = \frac{\textrm{exp}(e^{t,k}_{ij})}{\sum_{j \in \mathcal{N}(i)}{\textrm{exp}(e^{t,k}_{ij})}}\\
    M_{\textrm{atn}}^k(\bm{v}^t_i,\bm{v}^t_j; \bm{W}) = \alpha^{t,k}_{ij} \mathbf{W}^{v,k} \bm{v}^t_j,\\
    \bm{m}_i^t =  \bm{v}^t_i 
    + \Bigg{(}
    \Bigg\rVert^{K}_{k=1}
    \Bigg{[}\sum_{j \in \mathcal{N}(i)}
    M_{\textrm{atn}}^k(\bm{v}^t_i,\bm{v}^t_j; \bm{W}) \Bigg{]}\Bigg{)}\mathbf{W}^c
\end{gather}

Matrices $\mathbf{W}_{\cdot}^q,\mathbf{W}_{\cdot}^u,\mathbf{W}_{\cdot}^v,\mathbf{W}_{\cdot}^{r},\mathbf{W}_{\cdot}^{b},\mathbf{W}_{\cdot}^c$, are not shared across time steps (but are shared across nodes).

\vspace{1mm}
\noindent \textbf{Label Embedding Weight Sharing.\,} To enforce each label's input embedding to correspond to that particular label, the label embedding matrix weights $\mathbf{W}^{y}$ are shared with the readout projection matrix $\mathbf{W}^{o}$. In other words, $\mathbf{W}^{y}$ is used to produce the initial node vectors for \gyy{}, and then is used again to calculate the pre-sigmoid output values for each label, so $\mathbf{W}^{o}\equiv \mathbf{W}^y$.  This was shown beneficial in Seq2Seq models for machine translation  \cite{press2016using}.

\subsection{Loss Function}
The final output of \modelname{} networks $\hat{\bm{y}}$ are trained using the mean binary cross entropy (BCE) over all outputs $y_i$. For one sample, given true binary label vector $\bm{y}$ and predicted labels $\hat{\bm{y}}$, the output loss $\mathcal{L}_{out}$ is:
\vspace{-1pt}
\begin{equation}
    \label{eq:bce_loss}
    \mathcal{L}_{out}(\bm{y},\hat{\bm{y}}) = \frac{1}{L}\sum_{i=1}^{L} -(y_i\log(\hat{y}_i)+(1-y_i)\log(1-\hat{y}_i)) 
\end{equation}
\vspace{-2pt}

The final outputs $\hat{y}_i$ are computed from the final label node states $\bm{u}_i^T$ (Eq. \ref{eq:readout}). However, since \modelname{} networks iteratively update the label nodes from $t=0$ to $T$, we can ``probe'' the label nodes at each intermediate state from $t$=$1$ to $T$-$1$  and enforce an auxilary loss on those states. To do this, we use the same matrix $W^o$ to extract the intermediate prediction $\hat{y}^t_i$ at state $t$: 
$\hat{y}^t_i = R({\bm{u}}^t_i;{\mathbf{W}^o})$. We use the same BCE loss on the these predictions to compute intermediate loss $\mathcal{L}_{int}$:
\vspace{-2pt}
\begin{equation}
    \mathcal{L}_{int}(\bm{y},\hat{\bm{y}}^t) = \frac{1}{L} \sum_{i=1}^{L} -(y_i\log(\hat{y}^t_i)+(1-y_i)\log(1-\hat{y}^t_i)).
\end{equation}
\vspace{-6pt}

We note that the intermediate predictions $\hat{y}^t_i$ are computed for both $\bm{u}_i^t$ (after Label-to-Label message passing), as well as $\bm{u}_i^{t'}$ (after Feature-to-Label message passing). The final loss is a combination of both the original and intermediate, where the intermediate loss is weighted by $\lambda$:
\vspace{-2pt}
\begin{equation}
    \label{eq:lamp_loss}
    \mathcal{L}_{LaMP} = \mathcal{L}_{out}(\bm{y},\hat{\bm{y}}) + \lambda \sum_{t=1}^{T-1} \mathcal{L}_{int}(\bm{y},\hat{\bm{y}}^t)
\end{equation}
\vspace{-6pt}

In \modelname{} networks, $p(y_i|\{y_{j \neq i}\},\bm{z}_{1:S};\mathbf{W})$ is approximated by jointly representing $\{y_{1:L}\}$ using message passing from $\{\bm{z}_{1:S}\}$ and from the embeddings of all neighboring labels $\{y_{j \in \mathcal{N}(i)}\}$.
\subsection{\modelname{} Variation: Input Encoding with Feature Message Passing (FMP)}
\label{sec:encoder}

Thus far, we have assumed that we use the raw feature embeddings $\{\bm{z}_1,\bm{z}_2,...,\bm{z}_S\}$ to pass messages to the labels. However, we could also update the feature embeddings before they are passed to the label nodes by modelling the interactions between features. 

For a particular input $\bm{x}$, we first assume that the input features $\{x_{1:S}\}$ are nodes on a  graph, \gxx{}. \gxx{} $ = (V_{\textrm{xx}},E_{\textrm{xx}})$, $V_{\textrm{xx}}=\{x_{1:S}\}$, and $E$ includes all undirected pairwise edges connecting node $\bm{x}_i$ and node $\bm{x}_j$. \mpnnxx{}, parameterized by $\mathbf{W}_{\textrm{xx}}$, is used to pass messages between the input embeddings in order to update their states.  Nodes on \gxx{} are represented as embedding vectors $\{\bm{z}^t_1,\bm{z}^t_2,...,\bm{z}^t_S\}$, where the initial states $\{\bm{z}^0_{1:S}\}$ are obtained using embedding matrix $\mathbf{W}^x$ on input components $\{{x}_1,{x}_2,...,{x}_S\}$. The embeddings are then updated by \mpnnxx{} using message passing for $T$ time steps to produce $\{\bm{z}^T_1,\bm{z}^T_2,...,\bm{z}^T_S\}$.

To update input embedding $\bm{z}^t_i$, \mpnnxx{} uses attention message function $M_{\textrm{atn}}^t$ (Eq. \ref{eq:m_attention}) on all neighboring input embeddings $\{\bm{z}^t_{j \in \mathcal{N}(i)}\}$ to produce messages $\bm{m}^t_i$, and MLP update function $U_{\textrm{mlp}}$ (Eq. \ref{eq:u_attention}) to produce updated embedding $\bm{z}^{t+1}_i$:
\begin{gather}
    \bm{m}^t_i = \bm{z}^t_i + 
    \sum_{j \in \mathcal{N}(i)} M_{\textrm{atn}}(\bm{z}^t_i,\bm{z}^t_j; \bm{W}_{\textrm{xx}}), \\
    {\bm{z}}^{t+1}_i = \bm{m}^t_i + U_{\textrm{mlp}}(\bm{m}^{t}_i; \bm{W}_{\textrm{xx}}).
\end{gather}
If there exists a known \gxx{} graph, message $\bm{m}^t_i$ for node $i$ is computed using its neighboring nodes $j \in \mathcal{N}(i)$, where the neighbors $\mathcal{N}(i)$ are defined by the graph. If there is no known graph, we assume a fully connected \gxx{} graph, which means $\mathcal{N}(i) = \{j \in V_{\textrm{xx}}\}$. Inputs with a sequential ordering can be modelled as a fully connected graph using positional embeddings \cite{battaglia2018relational}.

In summary, \mpnnxx{} is used to update input feature nodes $\{\bm{z}^t_{1:S}\}$ by passing messages within the feature-interaction graph.  \mpnnxy{}, is used to update output label nodes $\{\bm{u}^t_{1:L}\}$ by passing messages from the features to labels (from input nodes $\{\bm{z}^t_{1:S}\}$ to output nodes $\{\bm{u}^t_{1:L}\}$).  \mpnnyy{}, is used to update output label nodes $\{\bm{u}^t_{1:L}\}$ by passing messages within the label-interaction graph (between label nodes). Once messages have been passed to update the feature and label nodes for $T$ integrative updates, a readout function $R$ is then used on the label nodes to make a binary classification prediction on each label, $\{\hat{y}_1,\hat{y}_2,...,\hat{y}_L\}$. Figure \ref{fig:model} shows the \modelname{} network without the feature-interaction graph.

\subsection{Advantages of \modelname{} Models}

\textbf{Efficiently Handling Dense Label Predictions.\,} It is known that autoregressive models such as RNN Seq2Seq suffer from the propagation of errors over the sequential positive label predictions. This makes it difficult for these models to handle dense, or many positive label, samples. In addition, autoregressive models require a time consuming post-processing step such as beam search to obtain the optimal label set. Lastly, autoregressive models require a predefined label ordering for training the sequential prediction, which can lead to instabilities at testing time \cite{vinyals2015order}.

Motivated by the drawbacks of autoregressive models for MLC, the proposed \modelname{} model removes the reliance on sequential predictions, beam search, and a chosen label ordering, while still modelling the label dependencies. This is particularly beneficial when the number of positive output labels is large (i.e. dense). \modelname{} networks predict the output \textit{set} of labels all at once, which is made possible by the fact that inference doesn't use a probabilistic chain, but there is still a representation of label dependencies via label to label attention. As an additional benefit, as noted by \cite{belanger2016structured}, it may be useful to maintain `soft' predictions for each label in MLC. This is a major drawback of the PCC models which make `hard' predictions of the positive labels, defaulting all other labels to 0. 

\vspace{4pt}
\noindent \textbf{Structure Agnostic.\,}  Many input or output types are instances 
where the relational structure is not made explicit, and must be inferred or assumed
\cite{battaglia2018relational}. \modelname{} networks allow for greater flexibility of both input structures (known structure such as sequence or graph, or unknown such as tabular), as well as output structures (e.g., known graph vs unknown structure). To the best of our knowledge, this is the first work to use MPNNs to \emph{infer} the relational structure of the data by using attention mechanisms.

\vspace{4pt}
\noindent \textbf{Interpretability.\,} Our formulation of \modelname{} allows us to visualize predictions in several different ways. First, since predictions are made in an iterative manner via graph update steps, we can ``probe'' each label's state at each step to get intermediate predictions. Second, we can visualize the attention weights which automatically learn the relational structure. Combining these two visualization methods allows us to see how the predictions change from the initial predictions given only the input sequence to the final state where messages have been passed from other labels, leading us to better insights for specific MLC samples.

\subsection{Connecting to Related Topics}
\label{sec:connecting}

\noindent \textbf{Structured Output Predictions.\, } The use of graph attention in \modelname{} models is closely connected to the literature of structured output prediction for MLC. \cite{ghamrawi2005collective} used conditional random fields (CRFs) \cite{lafferty2001conditional} to model dependencies among labels and features for MLC by learning a distribution over pairs of labels to input features, but these are limited to pairwise dependencies.

To overcome the naive pairwise dependency constraint of CRFs, structured prediction energy networks (SPENS) \cite{belanger2016structured} and related methods \cite{tu2018learning,gygli2017deep} locally optimize an unconstrained structured output. In contrast to SPENs which use an iterative refinement of the output label predictions, our method is a simpler feed forward block to make predictions in one step, yet still models dependencies through attention mechanisms on embeddings, which gives the added interpretability benefit.

\vspace{5pt}
\noindent \textbf{Multi-label Classification By Modeling Label Interaction Graphs.\,}
\cite{guo2011multi} formulate MLC using a label graph and they introduced a conditional dependency SVM where they first trained separate classifiers for each label given the input and all other true labels and used Gibbs sampling to find the optimal label set. The main drawback is that this method does not scale to a large number of labels. 
\cite{su2013multilabel} proposes a method to label the pairwise edges of randomly generated label graphs, and requires some chosen aggregation method over all random graphs. The authors introduce the idea that variation in the graph structure shifts the inductive bias of the base learners. Our fully connected label graph with attention on the neighboring nodes can be regarded as a form of graph ensemble learning \cite{hara2016analysis}. \cite{do2018attentional} use graph neural networks for MLC, but focus on graph inputs. They do not explicitly model label the label-to-label dependencies, thus resulting in a worse performance than \modelname{}.


\vspace{5pt}
\noindent \textbf{Graph Neural Networks (GNNs).\,} Passing embedding messages from node to neighbor nodes connects to a large body of literature on graph neural networks \cite{battaglia2018relational} and embedding models for structures \cite{dai2016discriminative}.

The key idea is that instead of conducting probabilistic operations (e.g., product or re-normalization), the proposed models perform nonlinear function mappings in each step to learn feature representations of structured components.  
\cite{gilmer2017neural,velivckovic2017graph,battaglia2016interaction} all follow similar ideas to pass the embedding from node to neighbor nodes or neighbor edges.

There have been many recent works extending the basic GNN framework to update nodes using various message passing, update, and readout functions 
\cite{kipf2016semi,hamilton2017representation,li2015gated,kearnes2016molecular,zheng2018unsupervised,battaglia2016interaction,gilmer2017neural,duvenaud2015convolutional}.
We refer the readers to \cite{battaglia2018relational} for a survey. However, none of these have used GNNs for MLC. In addition, none of these have attempted to learn the graph structure by using neural attention on fully connected graphs.

\section{Experiments}
\label{sec:experiments}

\begin{table}[t]
\scriptsize
\centering
\setlength{\tabcolsep}{2pt}
\caption{\small \textbf{ebF1 Scores} across all 8 datasets}
\label{ebf1_results}
\begin{tabular}{l|c|c|c|c|c|c|c|c}
 & \multicolumn{1}{l|}{Reuters} & \multicolumn{1}{l|}{Bibtex} & \multicolumn{1}{l|}{Bookmarks} & \multicolumn{1}{l|}{Delicious} & \multicolumn{1}{l|}{RCV1} & \multicolumn{1}{l|}{TFBS} & \multicolumn{1}{l|}{SIDER} & \multicolumn{1}{l}{NUSWIDE} \\ \hline
FastXML \cite{prabhu2014fastxml} & - & - & - & - & 0.841 & - & - & - \\
Madjarov \cite{madjarov2012extensive} & - & 0.434 & 0.257 & 0.343 & - & - & - & - \\
SPEN \cite{belanger2016structured} & - & 0.422 & 0.344 & 0.375 & - & - & - & - \\
RNN Seq2Seq \cite{nam2017maximizing} & 0.894 & 0.393 & 0.362 & 0.320 & \textbf{0.890} & 0.249 & 0.356 & 0.329 \\
Emb + MLP & 0.854 & 0.363 & 0.368 & 0.371 & 0.865 & 0.167 & \textbf{0.766}  & 0.371 \\ \hline
Emb + LaMP$_{el}$ & 0.859 & 0.379 & 0.351 & 0.358 & 0.868 & 0.289 & 0.767 & \textbf{0.376} \\
Emb + LaMP$_{fc}$ & 0.896 & 0.427 & 0.376 & 0.368 & 0.871 & 0.319 & 0.763 & \textbf{0.376} \\
Emb + LaMP$_{pr}$ & 0.895 & 0.424 & 0.373 & 0.366 & 0.870 & 0.317 & 0.765  & 0.372 \\ \hline
FMP + LaMP$_{el}$ & 0.883 & 0.435 & 0.375 & 0.369 & 0.887 & 0.310 & \textbf{0.766} & - \\
FMP + LaMP$_{fc}$ & \textbf{0.906} & 0.445 & \textbf{0.389} & \textbf{0.372} & 0.889 & \textbf{0.321} & 0.764 & - \\ 
FMP + LaMP$_{pr}$ & 0.902 & \textbf{0.447} & 0.386 & \textbf{0.372} & 0.887 & 0.321 & \textbf{0.766}& - 
\end{tabular}
\vspace{-5pt}
\end{table}

We validate our model on eight real world MLC datasets. These datasets vary in the number of samples, number of labels, input type (sequential, tabular, graph, vector), and output type (unknown, known label graph). They also cover a wide spectrum of input data types, including: raw English text (sequential form), binary word vector (tabular form), drug molecules (graph form), and images (vector form).  Data statistics are in Table \ref{table:datasets} and Section~\ref{sec:datasets}.  Due to the space limit, we move the details of evaluation metrics to Section~\ref{sec:metrics} and the hyper-parameters to Section~\ref{sec:hyper}. Details of  previous results from the state-of-the-art baselines are in Section~\ref{sec:baselines}.

\subsection{\modelname{} Variations} 
For \modelname{} models, we use two variations of input features, and three variations of Label-to-Label Message Passing. For input features, we use (1) \textbf{Emb}, which is the raw learned feature embeddings of dimension $d$, and (2) \textbf{FMP}\footnote{For NUS-WIDE, since we use the 128-dimensional cVLAD features as input to compare to \cite{do2018attentional}, we cannot use the \textbf{FMP} method.} which is the updated state of each feature embedding after 2 layers of Feature Message Passing, as explained in \ref{sec:encoder}. For each of the two input feature variations, we use three variations of the label graph which Label-to-Label Message Passing uses to update the labels given the input features, explained as follows.

\vspace{2pt}
\noindent \textbf{\modelname{}$_{el}$} uses an edgeless label graph and messages are not passed between labels, assuming no label dependencies.

\vspace{2pt}
\noindent \textbf{\modelname{}$_{fc}$} uses a fully connected label graph where each label is able to attend to all other labels (including itself) in order to compute the messages. 

\vspace{2pt}
\noindent \textbf{\modelname{}$_{pr}$} uses a prior label graph where each label is able to attend to only other labels from the known label graph (including itself) in order to compute the messages. For RCV1, we use the known tree label structure, and for TFBS we use known protein-protein interactions (PPI) from \cite{szklarczyk2016string}. For all other datasets, we create a graph where we place an edge on the adjacency matrix for all labels that co-occur in any sample for the training set. This is summarized in the last column of Appendix Table \ref{tab:datasets}.

\subsection{Performance Evaluation}
\label{sec:acc}

\textbf{ebF1.\,} Table \ref{ebf1_results} shows the most commonly used evaluation, example-based F1 (ebF1) scores, for the seven datasets. \modelname{} outperforms the baseline MLP models which assume no label dependencies, as well as RNN Seq2Seq, which models label dependencies using a classifier chain. More importantly, we compare using an output graph with no edges (\modelname{}$_{el}$), which assumes no label dependencies vs. an output graph with edges (\modelname{}$_{fc}$). The two models have the same architecture and number of parameters, with the only thing varying being the message passing between label nodes. We can see that for most datasets, modelling label dependencies using \modelname{}$_{fc}$ does in fact help. We found that using a known prior label structure (\modelname{}$_{pr}$) did not improve the results significantly.  \modelname{}$_{fc}$ predictions produced an average 1.8$\%$ ebF1 score increase over the independent \modelname{}$_{el}$ predictions. \modelname{}$_{pr}$ resulted in an average 1.7$\%$ ebF1 score increase over \modelname{}$_{el}$. When comparing to the MLP baseline, \modelname{}$_{fc}$ and \modelname{}$_{pr}$ produced an average 18.5$\%$ and 18.4$\%$ increase, respectively.

\vspace{4pt}
\noindent \textbf{miF1.\,} While high ebF1 scores indicate strong average F1 scores over all samples, the label-based Micro-averaged F1 (miF1) scores indicate strong results on the most frequent labels. Table \ref{mif1_results} shows the miF1 scores, for the all datasets. \modelname{}$_{fc}$ produced an average 1.6$\%$ miF1 score increase over the independent \modelname{}$_{el}$. \modelname{}$_{pr}$ produced an average 1.8$\%$ miF1 score increase over \modelname{}$_{el}$. When comparing to the MLP baseline, \modelname{}$_{fc}$ and \modelname{}$_{pr}$ resulted in an average 20.2$\%$ and 20.5$\%$ increase, respectively.

\vspace{4pt}
\noindent \textbf{maF1.\,} Contrarily, high label-based Macro-averaged F1 (maF1) scores indicate strong results on less frequent labels. Table \ref{mif1_results} shows maF1 scores, which show the strongest improvement of \modelname{}$_{fc}$ and \modelname{}$_{pr}$ variation over independent predictions. \modelname{}$_{fc}$ resulted in an average 2.4$\%$ maF1 score increase over the independent \modelname{}$_{el}$. \modelname{}$_{pr}$ produced an average 2.1$\%$ maF1 score increase over \modelname{}$_{el}$. This indicates that Label-to-Label message passing can help boost the accuracy of rare label predictions. When comparing to Emb + MLP, \modelname{}$_{fc}$ and \modelname{}$_{pr}$ produced an average 57.0$\%$ and 56.7$\%$ increase, respectively. 

\begin{table}[t]
\scriptsize
\centering
\setlength{\tabcolsep}{2pt}
\caption{\small \textbf{miF1 Scores} across all 8 datasets}
\label{mif1_results}
\begin{tabular}{l|c|c|c|c|c|c|c|c}
 & \multicolumn{1}{l|}{Reuters} & \multicolumn{1}{l|}{Bibtex} & \multicolumn{1}{l|}{Bookmarks} & \multicolumn{1}{l|}{Delicious} & \multicolumn{1}{l|}{RCV1} & \multicolumn{1}{l|}{TFBS} & \multicolumn{1}{l|}{SIDER} & \multicolumn{1}{l}{NUSWIDE} \\ \hline
FastXML \cite{prabhu2014fastxml} & - & - & - & - & 0.847 & - & - & - \\
SVM \cite{debole2005analysis} & 0.787 & - & - & - & - & - & - & - \\
GAML \cite{do2018attentional} & - & - & 0.333 & - & - & - & - & 0.398 \\
Madjarov \cite{madjarov2012extensive} & - & 0.462 & 0.268 & 0.339 & - & - & - & - \\
RNN Seq2Seq \cite{nam2017maximizing} & 0.858 & 0.384 & 0.329 & 0.329 & \textbf{0.884} & 0.311 & 0.389 & 0.418 \\
Emb + MLP & 0.835 & 0.389 & 0.349 & 0.385 & 0.855 & 0.218 & 0.795  & 0.465 \\ \hline
Emb + LaMP$_{el}$ & 0.842 & 0.413 & 0.334 & 0.372 & 0.858 & 0.401 & 0.797 & \textbf{0.472} \\
Emb + LaMP$_{fc}$ & 0.871 & 0.458 & 0.363 & 0.379 & 0.859 & 0.449 & 0.797 & 0.470 \\
Emb + LaMP$_{pr}$ & 0.877 & 0.462 & 0.363 & 0.380 & 0.859 & 0.448 & \textbf{0.798}  & 0.468 \\ \hline
FMP + LaMP$_{el}$ & 0.870 & 0.455 & 0.355 & 0.381 & 0.877 & 0.445 & 0.795 & - \\
FMP + LaMP$_{fc}$ & 0.886 & 0.465 & \textbf{0.373} & 0.384 & 0.877 & \textbf{0.450} & 0.795 & - \\
FMP + LaMP$_{pr}$ & \textbf{0.889} & \textbf{0.473} & 0.371 & \textbf{0.386} & 0.877 & 0.449 & 0.797& - 
\end{tabular}
\vspace{-5pt}
\end{table}

\vspace{4pt}
\noindent \textbf{Other Metrics.\,} Due to space constraints, we report subset accuracy in Appendix (supplementary) Table \ref{tab:acc_results}. RNN Seq2Seq models mostly perform all other models for this metric since they are trained to maximize it\cite{nam2017maximizing}. However, for all other metrics, RNN Seq2Seq does not perform as well, concluding that for most applications, PCC models aren't necessary. We also report Hamming Accuracy in Appendix Table \ref{tab:ha_results}, and we note that \modelname{} networks outperform or perform similarly to baseline methods, but we observe that this metric is mostly unhelpful. 

\vspace{4pt}
\noindent \textbf{Metrics Performance Summary.\,} While \modelname{} does not explicitly model label dependencies as autoregressive or structured prediction models do, the attention weights do learn some dependencies among labels (Section \ref{sec:interpret}). This is indicated by the fact that \modelname{}, which uses Label-to-Label attention, mostly outperforms the ones which don't, indicating that it is learning label dependencies. 

\vspace{4pt}
\noindent \textbf{Speed.\,} \modelname{} results in a mean of 1.7x and 5.0x training and testing speedups, respectively, over the previous state-of-the-art probabilistic MLC method, RNN Seq2Seq. Speedups over RNN Seq2Seq model are shown in Table \ref{tab:speed}.

\begin{table}[t]
\scriptsize
\centering
\setlength{\tabcolsep}{2pt}
\caption{\footnotesize \textbf{maF1 Scores} across all 8 datasets}
\label{maf1_results}
\begin{tabular}{l|c|c|c|c|c|c|c|c}
 & \multicolumn{1}{l|}{Reuters} & \multicolumn{1}{l|}{Bibtex} & \multicolumn{1}{l|}{Bookmarks} & \multicolumn{1}{l|}{Delicious} & \multicolumn{1}{l|}{RCV1} & \multicolumn{1}{l|}{TFBS} & \multicolumn{1}{l|}{SIDER} & \multicolumn{1}{l}{NUSWIDE} \\ \hline
SVM \cite{debole2005analysis} & 0.468 & - & - & - & - & - & - & - \\
FastXML \cite{prabhu2014fastxml} & - & - & - & - & 0.592 & - & - & - \\
GAML \cite{do2018attentional} & - & - & 0.217 & - & - & - & - & 0.114 \\
Madjarov \cite{madjarov2012extensive} & - & 0.316 & 0.119 & 0.142 & - & - & - & - \\
RNN Seq2Seq \cite{nam2017maximizing} & 0.457 & 0.282 & 0.237 & 0.166 & 0.741 & 0.210 & 0.207 & 0.143 \\
Emb + MLP & 0.366 & 0.275 & 0.248 & 0.180 & 0.667 & 0.094 & 0.665  & 0.173 \\ \hline
Emb + LaMP$_{el}$ & 0.476 & 0.308 & 0.229 & 0.176 & 0.680 & 0.326 & 0.666 & 0.198 \\
Emb + LaMP$_{fc}$ & 0.547 & 0.366 & 0.271 & 0.192 & 0.691 & 0.362 & 0.663 & \textbf{0.203} \\
Emb + LaMP$_{pr}$ & \textbf{0.560} & 0.372 & 0.267 & 0.192 & 0.698 & 0.365 & 0.663 & 0.196 \\ \hline
FMP + LaMP$_{el}$ & 0.508 & 0.353 & 0.266 & 0.192 & 0.742 & \textbf{0.368} & 0.664 & - \\
FMP + LaMP$_{fc}$ & 0.520 & 0.371 & \textbf{0.286} & 0.195 & \textbf{0.743} & 0.364 & \textbf{0.668} & - \\
FMP + LaMP$_{pr}$ & 0.517 & \textbf{0.376} & 0.280 & \textbf{0.196} & 0.740 & 0.364 & 0.664& -\\\hline
\end{tabular}
\vspace{-5pt}
\end{table}

\subsection{Interpretability Evaluation} 
\label{sec:interpret}
The structure of \modelname{} networks allows for three different types of visualization methods to understand how the network predicts each label. We explain the three types here and show the results for a sample from the Bookmarks dataset using the FMP + \modelname{}$_{fc}$ model.

\vspace{4pt}
\noindent \textbf{Intermediate Output Prediction.\,} One advantage of the multi step formulation of label embedding updates is that it gives us the ability to probe the state of each label at intermediate steps and view the model's predictions at those steps. To do this, we use the readout function $R$ on each intermediate label embeddings state $\bm{u}_i^t$ to find the probability that the label embedding would predict a positive label. In other words, this is the post-sigmoid output of the readout function of each embedding $R(\bm{u}_i^t;\bf{W}^o)$ at each step $t=1,...,T$. We note that each step contains two stages: $t.1$ is the output after the Feature-to-Label message passing, and $t.2$ is output after the Label-to-Label message passing. The output after the second stage of the final step (i.e. $T.2$) is the model's final output.

\begin{figure*}[t]
    \centering
    \includegraphics[width=1.0\linewidth]{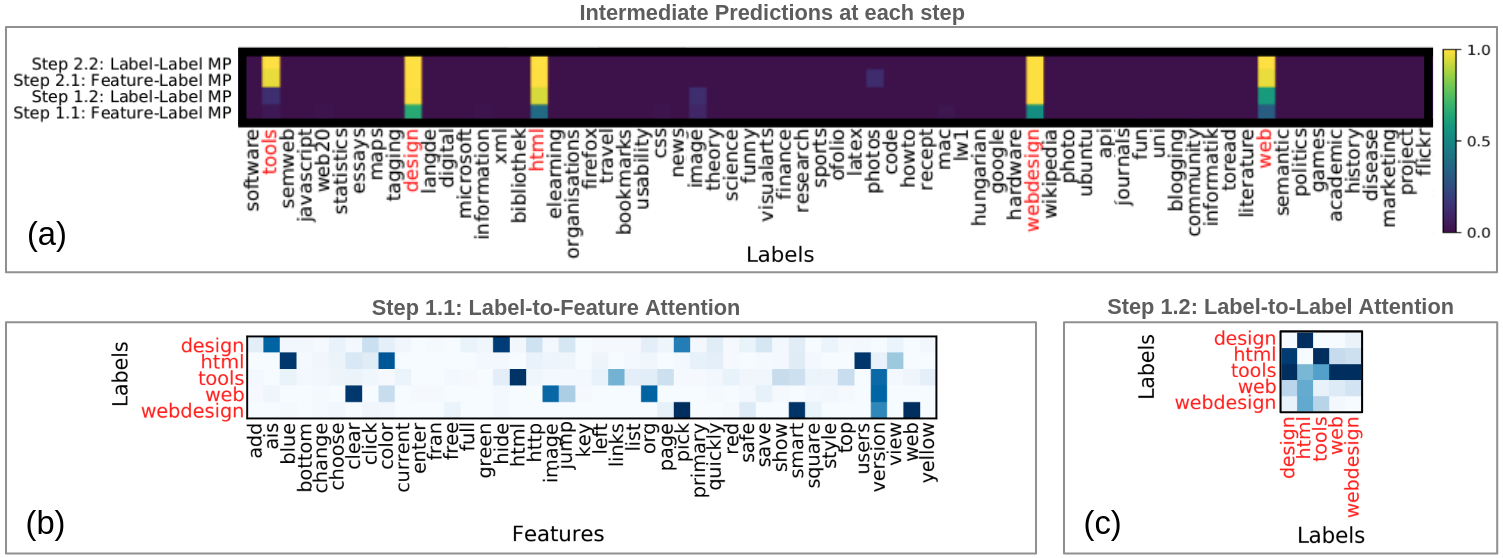}
     \caption{
     \footnotesize \textbf{(a) Visualization of Model Predictions and Attention Weights}  Intermediate Predictions: this shows the readout function predictions for each intermediate state in the two update steps. \textbf{(b) Label-to-Feature Attention Weights} for the first step of Feature-to-Label message passing ($1.1$). \textbf{(c) Label-to-Label Attention Weights} for the first step of Label-to-Label message passing ($1.2$).
     }
    \label{fig:visualization}
\end{figure*}

Figure~\ref{fig:visualization} (a.) shows the intermediate prediction outputs from the $T=2$ step model. On the horizontal axis are a selected subset of all possible labels, with the red colored axis labels being all true positive labels. On the vertical axis, each row represents one of the label embedding states in the $T=2$ step model. Each cell represents the readout function's prediction for each label embedding's state. The brighter the grid cell, the more likely that label is positive at the current stage. Starting from the bottom, the first row $(1.1)$ shows the prediction of each label after the first Feature-to-Label message passing. The second row ($1.2$) shows the prediction of each label after the first Label-to-Label message passing.  This is then repeated once more in $(2.1)$ and $(2.2)$ for the second layer's output states, where the final output, $2.2$ is the network's final output predictions.
The most important aspect of this figure is that we can see the labels ``design'', ``html'', and ``web design'', all change from weakly positive to strongly positive after the first Label-to-Label message passing step (row $1.2$). In other words, this indicates that these labels change to a strongly positive prediction by passing messages between each other.

\vspace{2pt}
\noindent \textbf{Label-to-Feature Attention.\,}
While the iterative prediction visualization shows how the model updates its prediction of each label, it doesn't explicitly show how or why. To understand why each label changes its predictions, we first look at the Feature-to-Label attention, which tells us the input nodes that each label node attends to in order to update its state (and thus producing the predictions in Figure~\ref{fig:visualization} (a.)).  Figure~\ref{fig:visualization} (b.) shows us which input nodes (i.e. features) each of the positive label attends to in order to make its first update step $1.1$. The colors represent the post-softmax attention weight (summed over the 4 attention heads), with the darker cells representing high attention. In this example, we can see that the ``web design'' label attends to the ``pick'', ``smart'', and ``version'' features, but as we can see from the first row of  Figure~\ref{fig:visualization} (a.), prediction for the current state of the ``web design'' label isn't very strong yet.

\vspace{2pt}
\noindent \textbf{Label-to-Label Attention.\,}
Label-to-Feature attention shows us the input nodes that each label node attends to in order to make its first update, but the second step of the label graph update is the Label-to-Label message passing step where labels are updated according to the states of all other nodes after the first Feature-to-Label message passing. Figure~\ref{fig:visualization} (c.) shows us the first Label-to-Label attention stage $1.2$ where each label node can attend to the other label nodes in order to update its state. Here we show only the Label-to-Label attention for the positive labels in this example. We then look at the second row of Figure~\ref{fig:visualization} (a.) which shows the model's prediction of each label node after the Label-to-Label attention. The interesting thing to note here is we can see many of the true positive labels change their state to positive after the positive labels attend to each other during the Label-to-Label attention step, indicating that dependencies are learned.

\vspace{2pt}
Attention weights for the second step $t=2$ are not as interpretable since they model higher order interactions. We have added these plots in Appendix Fig. \ref{fig:viz2}.


\begin{SCtable}
\scriptsize
\centering
\setlength{\tabcolsep}{2pt}
\begin{tabular}{l|l|l}
Dataset   & Training      & Testing      \\ \hline
Reuters   & 0.788 (1.5x)  & 0.116 (2.1x) \\
Bibtex    & 0.376 (2.1x)  & 0.080 (2.1x) \\
Delicious & 3.172 (1.1x)  & 0.473 (3.2x) \\
Bookmarks & 9.664 (1.2x)  & 1.849 (1.3x) \\
RCV1      & 98.346 (1.2x) & 1.003 (1.7x) \\
TFBS      & 187.14 (2.5x) & 13.04 (4.2x) \\
NUS-WIDE  & 3.201 (1.2x)  & 0.921 (8.0x) \\
SIDER     & 0.027 (2.5x)  & 0.003 (21x) 
\end{tabular}
\label{tab:speed}
\caption{\scriptsize \textbf{Speed.} Each column shows training or testing speed for \modelname{} in minutes per epoch. Speedups over RNN Seq2Seq are in parentheses. Since \modelname{} does not depend on sequential prediction, we see a drastic speedup, especially during testing where RNN Seq2Seq requires beam search.}
\end{SCtable}
\vspace{-10pt}

\section{Conclusion}

In this work we present \fullmodelname{} (\modelname{}) Networks which achieve better than, or close to the same accuracy as previous methods across five metrics and seven datasets. In addition, the iterative label embedding updates with attention of \modelname{} provide a straightforward way to shed light on the model's predictions and allow us to extract three forms of visualizations, including conditional label dependencies which influence MLC classifications.


 \bibliography{nips2018} 
 \bibliographystyle{splncs04}

\clearpage
\newpage


\section{Appendix: MLC Background}
\label{sec:background}
\subsection{Background of Multi-Label Classification: } 
\label{sec:MLC}
MLC has a rich history in text \cite{mccallum1999multi,ueda2003parametric}, images \cite{tsoumakas2006multi,elisseeff2002kernel}, bioinformatics \cite{tsoumakas2006multi,elisseeff2002kernel}, and many other domains. MLC methods can roughly be broken into several groups, which are explained as follows.

Label powerset models (LP) \cite{tsoumakas2007random,read2011classifier}, classify each input into one label combination from the set of all possible combinations $\mathcal{Y} = \{\{1\}, \{2\},...,\{1, 2,..., L\}\}$. LP explicitly models the joint distribution by predicting the one subset of all positive labels. Since the label set $Y$ grows exponentially in the number of total labels ($2^{L}$), classifying each possible label set is intractable for a modest $L$.  In addition, even in small $L$ tasks, LP suffers from the ``subset scarcity problem'' where only a small amount of the label subsets are seen during training, leading to bad generalization.

Binary relevance (BR) methods predict each label separately as a logistic regression classfier for each label \cite{zhang2005k,godbole2004discriminative}. The na\"ive approach to BR prediction is to predict all labels independently of one another, assuming no dependencies among labels. That is, BR uses the following conditional probability parameterized by learned weights $W$:
\vspace{-0.3em}
\begin{equation} \label{eq:1}
    P_{BR}(Y|X;W) = \prod_{i=1}^{L}p(Y_i|X_{1:S};W)
\end{equation}

Probabilistic classifier chain (PCC) methods \cite{dembczynski2010bayes,read2009classifier} are autoregressive models that estimate the true joint probability of output labels given the input by using the chain rule, predicting one label at a time:

\begin{equation} \label{eq:2}
    P_{PCC}(Y|X;W) = \prod_{i=1}^{L}p(Y_i|Y_{1:i-1},X_{1:S};W)
\end{equation}

Two issues with PCC models are that inference is very slow if $L$ is large, and the errors propagate as $L$ increases \cite{montanes2014dependent}. To mitigate the problems with both LP and PCC methods, one solution is to only predict the true labels in the LP subset. In other words, only predicting the positive labels (total of $\rho$ for a particular sample) and ignoring all other labels, which we call PCC$+$. Similar to PCC, the joint probability of PCC$+$ can be computed as product of conditional probabilities, but unlike PCC, only $\rho < L$ terms are predicted as positive:
\vspace{-0.3em}
\begin{equation} \label{eq:3}
    P_{PCC+}(Y|X;W) = \prod_{i=1}^{\rho}p(Y_{p_i}|Y_{p_{1:i-1}},x_{1:S};W)
\end{equation}
\vspace{-1em}

This can be beneficial when the number of possible labels $L$ is large, reducing the total number of prediction steps. 
However, in both PCC and PCC$+$, inference is done using beam search, which is a costly dynamic programming step to find the optimal prediction.

Recently, Recurrent neural network (RNN) based encoder-decoder models following PCC and PCC$+$ have shown state-of-the-art performance for solving MLC. However, the sequential nature of modeling label dependencies through an RNN limits its ability in parallel computation, predicting dense labels, and providing interpretable results.

The main drawback of classifier chain models is that their inherently sequential nature precludes parallelization during training and inference. This can be detrimental when there are a large number of positive labels as the classifier chain has to sequentially predict each label, and often requires beam search to obtain the optimal set. Aside from time-cost disadvantages, PCC methods have several other drawbacks. First, PCC methods require a defined ordering of labels for the sequential prediction, but MLC output labels are an unordered set, and the chosen order can lead to prediction instability \cite{nam2017maximizing}. Secondly, even if the optimal ordering is known, PCC methods struggle to accurately capture long-range dependencies among labels in cases where the number of positive labels is large (i.e., dense labels). For example, the Delicious dataset we used in the experiment has a median of 19 positive labels per sample, so it can be difficult to correctly predict the labels at the end of the prediction chain. Lastly,  many real-world applications prefer interpretable predictors. For instance, in the task of predicting which proteins (labels) will bind to a DNA sequence based binding site, users care about how a prediction is made and how the interactions among labels (proteins) influence the binding predictions. An important task is modelling what is known as ``co-binding'' effects, where one protein will \textit{only} bind if there is another specific protein already binding, or similarly will not bind if there is another already binding.

\modelname{} methods approximate the following factored formulation, where $\mathcal{N}(Y_i)$ denotes the neighboring nodes of $Y_i$.  
\begin{equation}
    \label{eq:factored_formula}
    P_{G2G}(Y|X;W) = \prod_{i=1}^{L}p(Y_i|\{Y_{\mathcal{N}(Y_i)}\},X_{1:S};W).
\end{equation}

\subsection{Seq2Seq Models}

In machine translation (MT), sequence-to-sequence (Seq2Seq) models have proven to be the superior method, where an encoder RNN reads the source language sentence into an encoder hidden state, and a decoder RNN translates the hidden state into a target sentence, predicting each word autoregressively \cite{sutskever2014sequence}. 
\cite{bahdanau2014neural} improved this model by introducing ``neural attention'' which allows the decoder RNN to ``attend'' to every encoder word at each step of the autoregressive translation. 

Recently, \cite{nam2017maximizing} showed that, across several metrics, state-of-the-art MLC results could be achieved by using a recurrent neural network (RNN) based encoder-to-decoder framework for Equation \ref{eq:3} (PCC$+$). They use a Seq2Seq RNN model (Seq2Seq Autoregressive) which uses one RNN to encode $\bm{x}$, and a second RNN to predict each positive label sequentially, until it predicts a `stop' signal. This type of model seeks to maximize the `subset accuracy', or correctly predict every label as its exact 0/1 value.

\cite{vaswani2017attention} eliminated the need for the recurrent network in MT by introducing the Transformer. Instead of using an RNN to model dependencies, the Transformer explicitly models pairwise dependencies among all of the features by using attention \cite{bahdanau2014neural,Xu2015} between signals. This speeds up training time because RNNs can't be fully parallelized but, the transformer still uses an autoregressive decoder.

\subsection{Drawbacks of Autoregressive Models}
\label{drawbacks}

Seq2Seq MLC \cite{nam2017maximizing} uses an encoder RNN encoding elements of an input sequence, a decoder RNN predicting output labels one after another, and beam search that computes the probability of the next $T$ predictions of labels and then chooses the solution with the max combined probability.  

Autoregressive models have been proven effective for machine translation and MLC \cite{sutskever2014sequence,bahdanau2014neural,nam2017maximizing}. However, predictions must be made sequentially, eliminating parallelization. Also, beam search is typically used at test time to find optimal predictions. But beam search is limited by the time cost of large beams sizes, making it difficult to optimally predict many output labels \cite{koehn2017six}.

In addition to speed constraints, beam search for autoregressive inference introduces a second drawback: initial wrong predictions will propagate when using a modest beam size (e.g. most models use a beam size of 5). This can lead to significant decreases in performance when the number of positive labels is large. For example, the Delicious dataset has a median of 19 positive labels per sample, and it can be very difficult to correctly predict the labels at the end of the prediction chain.

Autoregressive models are well suited for machine translation because these models mimic the sequential decoding process of real translation. However, for MLC, the output labels have no intrinsic ordering. While the joint probability of the output labels is independent of the label ordering via autoregressive based inference, the chosen ordering can make a difference in practice \cite{vinyals2015order,nam2017maximizing}. Some ordering of labels must be used during training, and this chosen ordering can lead to unstable predictions at test time.

Our \modelname{} connects to \cite{gu2017non} who removed the autoregressive decoder in MT with the Non-Autoregressive Transformer. In this model, the encoder makes a proxy prediction, called ``fertilities'', which are used by the decoder to predict all translated words at once. The difference between their model and ours is that we have a constant label at each position, so we don't need to marginalize over all possible labels at each position.

\section{Appendix: Dataset Details}

\begin{table*}[h] \footnotesize
\centering 
\scriptsize
\setlength{\tabcolsep}{0.3em}
\caption{\small \textbf{Dataset Statistics}. We use 7 well studied MLC datasets, plus our own TFBS protein binding dataset. Each dataset varies in the input type, number of samples, number of labels, and number of input features. The last column shows the prior graph structure type we explore for the \modelname{}$_{pr}$ model.} \label{table:datasets}
\begin{tabular}{l|l|l|l|l|l|l|l|l} 
Dataset  & Input Type & Domain & \#Train & \#Val & \#Test & \begin{tabular}[c]{@{}l@{}}Labels\\ ($L$)\end{tabular} & Features & \multicolumn{1}{l}{\begin{tabular}[c]{@{}l@{}}Prior\\ Graph\\ Structure\end{tabular}} \\ \hline 
Reuters-21578 & Sequential  & Text & 6,993 & 777 & 3,019 & 90 & 23,662 & Co-occur\\ 
RCV1-V2 & Sequential  & Text & 703,135 & 78,126 & 23,149 & 103 & 368,998 & Tree\\ 
TFBS & Sequential & Biology & 1,671,873 & 301,823 & 323,796 & 179 & 4 & PPI\\
BibTex  & Binary Vector & Text & 4,377 & 487 & 2,515 & 159 & 1,836 & Co-occur\\ 
Delicious  & Binary Vector & Text & 11,597 & 1,289 & 3,185 & 983 & 500 & Co-occur\\ 
Bookmarks & Binary Vector & Text & 48,000 & 12,000 & 27,856 & 208 & 368,998 & Co-occur\\ 
NUS-WIDE & Vector & Image & 129,431 & 32,358 & 107,859 & 85 & 128 & Co-occur\\
SIDER & Graph & Drug & 1,141 & 143 & 143 & 27 & 37 & Co-occur
\end{tabular} 
\label{tab:datasets}
\end{table*}

\begin{table*}[h] \footnotesize
\centering 
\scriptsize
\setlength{\tabcolsep}{0.3em}
\caption{\small \textbf{Additional Dataset Statistics} Here we show additional statistics of datasets with respect to the specific number of labels for each dataset. This shows how each dataset has a varying degree of MLC difficulty regarding the number of labels which need to be predicted. } \label{table:datasets}
\begin{tabular}{l|l|l|l|l|l|l} 
Dataset  & \multicolumn{1}{l|}{\begin{tabular}[c]{@{}l@{}}Mean\\ Labels\\ /Sample\end{tabular}} & \begin{tabular}[c]{@{}l@{}}Median \\ Labels\\ /Sample\end{tabular} & \begin{tabular}[c]{@{}l@{}}Max\\ Labels\\ /Sample\end{tabular} & \begin{tabular}[c]{@{}l@{}}Mean\\ Samples\\ /Label\end{tabular} & \begin{tabular}[c]{@{}l@{}}Median\\ Samples\\ /Label\end{tabular} & \begin{tabular}[c]{@{}l@{}}Max\\ Samples\\/Label\end{tabular} \\ \hline 
Reuters-21578 & 1.23& 1& 15 & 106.50 & 18 & 2,877 \\ 
RCV1-V2 & 3.21& 3& 17 & 24,362 & 7,250 & 363,991 \\ 
TFBS & 7.62 & 2& 178 & 84,047 & 45,389 & 466,876 \\ 
BibTex  & 2.38& 2& 28 & 72.79 & 54 & 689 \\ 
Delicious  & 19.06 & 20 & 25 & 250.15 & 85& 5,189 \\ 
Bookmarks & 2.03& 1& 44 & 584.67 & 381 & 4,642 \\ 
NUS-WIDE & 1.86 & 1 & 12 & 3721.7 & 1104 & 44255 \\ 
SIDER & 15.3 & 16 & 26 & 731.07 & 851 & 1185
\end{tabular} 
\label{tab:dataset_label_stats}
\end{table*}

\section{Appendix: Extra Metrics}
Here we provide the results from an extra two metrics: subset accuracy and hamming accuracy.

\begin{table}[H]
\scriptsize
\centering
\setlength{\tabcolsep}{2pt}
\caption{\small \textbf{Subset Accuracy Scores} across all 7 datasets}
\label{tab:acc_results}
\begin{tabular}{l|c|c|c|c|c|c|c|c}
 & \multicolumn{1}{l|}{Reuters} & \multicolumn{1}{l|}{Bibtex} & \multicolumn{1}{l|}{Bookmarks} & \multicolumn{1}{l|}{Delicious} & \multicolumn{1}{l|}{RCV1} & \multicolumn{1}{l|}{TFBS} & \multicolumn{1}{l|}{SIDER} & \multicolumn{1}{l}{NUSWIDE} \\ \hline
Madjarov & - & \textbf{0.202} & 0.209 & \textbf{0.018} & - & - & - & - \\
RNN Seq2Seq & \textbf{0.837} & 0.195 & \textbf{0.273} & 0.016 & \textbf{0.6798} & \textbf{0.114} & 0.000 & 0.252 \\
Emb + MLP & 0.774 & 0.151 & 0.234 & 0.180 & 0.620 & 0.040 & 0.014  & 0.263 \\ \hline
Emb + LaMP$_{el}$ & 0.757 & 0.141 & 0.214 & 0.176 & 0.619 & 0.077 & 0.014 & 0.268 \\
Emb + LaMP$_{fc}$ & 0.813 & 0.171 & 0.234 & 0.192 & 0.630 & 0.086 & 0.007 & \textbf{0.269} \\
Emb + LaMP$_{pr}$ & 0.813 & 0.169 & 0.232 & 0.192 & 0.621 & 0.087 & 0.007 & 0.267 \\ \hline
FMP + LaMP$_{el}$ & 0.808 & 0.158 & 0.231 & 0.192 & 0.656 & 0.084 & 0.007 & - \\
FMP + LaMP$_{fc}$ & 0.835 & 0.182 & 0.242 & 0.195 & 0.660 & 0.090 & 0.014 & - \\
FMP + LaMP$_{pr}$ & 0.828 & 0.185 & 0.241 & 0.196 & 0.659 & 0.090 & 0.007& - \\
\end{tabular}
\end{table}

\begin{table}[H]
\scriptsize
\centering
\setlength{\tabcolsep}{2pt}
\caption{\small \textbf{Hamming Accuracy} across all 7 datasets}
\label{tab:ha_results}
\begin{tabular}{l|c|c|c|c|c|c|c|c}
 & \multicolumn{1}{l|}{Reuters} & \multicolumn{1}{l|}{Bibtex} & \multicolumn{1}{l|}{Bookmarks} & \multicolumn{1}{l|}{Delicious} & \multicolumn{1}{l|}{RCV1} & \multicolumn{1}{l|}{TFBS} & \multicolumn{1}{l|}{SIDER} & \multicolumn{1}{l}{NUSWIDE} \\ \hline
Madjarov & - & 0.988 & 0.991 & 0.982 & - & - & - & - \\
RNN Seq2Seq & 0.996 & 0.985 & 0.990 & 0.980 & 0.9925 & 0.961 & 0.593 & 0.980 \\
Emb + MLP & 0.996 & 0.987 & 0.991 & 0.982 & 0.992 & 0.959 & 0.752 & 0.980 \\
Emb + LaMP$_{el}$ & 0.996 & 0.987 & 0.991 & 0.982 & 0.992 & 0.963 & 0.750  & 0.980 \\ \hline
Emb + LaMP$_{fc}$ & 0.997 & 0.988 & 0.992 & 0.982 & 0.992 & 0.964 & 0.752 & 0.980 \\
Emb + LaMP$_{pr}$ & 0.997 & 0.988 & 0.991 & 0.982 & 0.992 & 0.964 & 0.751  & 0.980 \\ \hline
FMP + LaMP$_{el}$ & 0.997 & 0.987 & 0.991 & 0.982 & 0.993 & 0.964 & 0.748 & - \\
FMP + LaMP$_{fc}$ & 0.997 & 0.988 & 0.992 & 0.982 & 0.993 & 0.964 & 0.749 & - \\
FMP + LaMP$_{pr}$ & 0.997 & 0.988 & 0.992 & 0.982 & 0.993 & 0.964 & 0.747& -
\end{tabular}
\end{table}

\section{Appendix: More about Experiments}

\begin{figure*}[ht]
    \centering
    \includegraphics[width=1.0\linewidth]{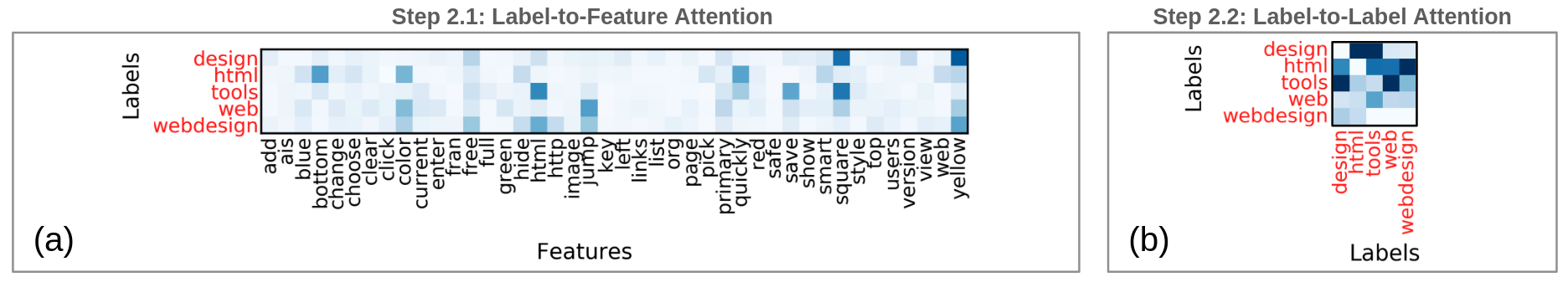}
     \caption{
     \footnotesize This shows the step $t=2$ visualization results from Fig \ref{fig:visualization} (a). \textbf{2.1 Label-to-Feature Attention Weights (b)}. \textbf{2.2 Label-to-Label Attention Weights (c)} 
     }
    \label{fig:viz2}
\end{figure*}

\subsection{Datasets}
\label{sec:datasets}

We test our method against baseline methods on seven different multi-label sequence classification datasets. The datasets are summarized in Table \ref{table:datasets}. We use Reuters-21578 \cite{lewis2004rcv1}, Bibtex \cite{tsoumakas2009mining}, Delicious \cite{tsoumakas2008effective}, Bookmarks \cite{katakis2008multilabel}, RCV1-V2 \cite{lewis2004rcv1}, our own DNA protein binding dataset (TFBS) from \cite{encode2012integrated}, and SIDER \cite{kuhn2015sider}, which is side effects of drug molecules. As shown in the table, each dataset has a varying number of samples, number of labels, positive labels per sample, and samples per label. For BibTex and Delicious, we use 10\% of the provided training set for validation. For the TFBS dataset, we use 1 layer of convolution at the first layer to extract ``words'' from the DNA characters (A,C,G,T), as commonly done in deep learning models for DNA. 

For datasets which have sequential ordering of the input components (Reuters, RCV1), we add a positional encoding to the word embedding as used in \cite{vaswani2017attention} (sine and cosine functions of different frequencies) to encode the location of each word in the sentence. For datasets with no ordering or graph stucture (Bibtex, Delicious, Bookmarks, which use bag-of-word input representations) we do not use positional encodings. For inputs with an explicit graph representation (SIDER), we use the known graph structer.

\subsection{Evaluation Metrics}
\label{sec:metrics}
Multi-label classification methods can be evaluated with many different metrics which each evaluate different strengths or weaknesses. We use the same 5 evaluation metrics from \cite{nam2017maximizing}.

All of our autoregressive models predict only the positive labels before outputting a stop signal. This is a special case of PCC models 
, which have been shown to outperform the binary prediction of each label in terms of performance and speed. These models use beam search at inference time with a beam size of 5. For the non-autoregressive models, to convert the labels to $\{0,1\}$ we chose the best threshold on the validation set from the same set of thresholds used in \cite{tu2018learning}.

\textbf{Example-based measures} are defined by comparing the target vector $\bm{y}$ to the prediction vector $\bm{\hat{y}}$.
    Subset Accuracy (ACC) requires an exact match of the predicted labels and the true labels: ACC$(\bm{y}, \bm{\hat{y}}) = \mathbb{I}[\bm{y} = \bm{\hat{y}}]$.
    Hamming Accuracy (HA) evaluates how many labels are correctly predicted in $\hat{y}$: HA$(\bm{y}, \bm{\hat{y}}) = \frac{1}{L} \sum_{j=1}^{L} \mathbb{I}[y_j = \hat{y}_j]$.
    Example-based F1 (ebF1) measures the ratio of correctly predicted labels to the sum of the total true and predicted labels: $\frac{2\sum_{j=1}^L y_j\hat{y}_j}{\sum_{j=1}^L y_j + \sum_{j=1}^L \hat{y}_j}$.

\textbf{Label-based measures} treat each label $y_j$ as a separate two-class prediction problem, and compute the number of true positives ($tp_j$), false positives ($fp_j$), and false negatives ($fn_j$) for a label. Macro-averaged F1 (maF1) measures the label-based F1 averaged over all labels: $\frac{1}{L}\sum_{j=1}^L \frac{2tp_j}{2tp_j + fp_j + fn_j}$.  Micro-averaged F1 (miF1) measures the label-based F1 averaged over each sample: $\frac{\sum_{j=1}^L 2tp_j}{\sum_{j=1}^L 2tp_j + fp_j + fn_j}$. High maF1 scores usually indicate high performance on less frequent labels. High miF1 scores usually indicate high performance on more frequent labels.

\FloatBarrier



\subsection{Model Hyperparameter Tuning}
\label{sec:hyper}

For all 7 datasets (Table \ref{table:datasets}), we use the same \modelname{} model with $T$=2 time steps, $d=512$, and $K$=4 attention heads. We trained our models on an NVIDIA TITAN X Pascal with a batch size of 32. We used Adam \cite{kingma2014adam} with betas$=(0.9, 0.999)$, eps$=$1e-08, and a learning rate of 0.0002 for each dataset. We used dropout of $p=0.2$ for the smaller datasets (Reuters, Bibtex, SIDER), and dropout of $p=0.1$ for all other datasets. The \modelname{} models also use layer normalization \cite{ba2016layer} around each of the attention and feedforward layers. All \modelname{} models are trained with the \modelname{} loss (Eq. \ref{eq:lamp_loss}). The hyperparameter $\lambda$ is selected from the best performing value in $\{0,0.1,0.2,0.3\}$ for each model. MLP models are trained with regular binary cross entropy (Eq. \ref{eq:bce_loss}), and RNN Seq2Seq model are trained with cross entropy across all possible labels at each position. To convert the soft predictions into $\{0,1\}$ values, we use the same thresholds in \cite{belanger2016structured} and select the best one for each metric on the validation set. For the TFBS dataset, which uses DNA input sequences, we use one layer of convolution to get 512 dimensional embeddings as commonly done for deep neural network prediction tasks on DNA sequences.

\subsection{Baseline Comparisons}
\label{sec:baselines}

Briefly, we compare against the following methods for all reported datasets and metrics. 
For those results named as \textrm{``Madjarov''}: we take the best method for each reported metric from \cite{madjarov2012extensive} who compared 12 different types of models including SVMs, decision trees, boosting, classification rules, and neural networks. For results of \textrm{``SPEN''}: Structured prediction energy networks from \cite{belanger2016structured}. For results of ``SVM'': SVM method from the Reuters dataset authors \cite{debole2005analysis}. For results of ``FastXML'': Fast random forest model   \cite{prabhu2014fastxml}. For results of ``GAML'': graph attention for MLC from \cite{do2018attentional}. For ``RNN Seq2Seq'': RNN Sequence to Sequence model from \cite{nam2014large} which is a PCC model that predicts only the positive labels. For ``Emb + MLP'': we use the mean embeddings of all input features as the input to a 4 layer multi-layer perceptron (MLP). This is a BR baseline which predicts all labels independently.

\end{document}